\documentclass[sigconf]{acmart}
\usepackage{fix-cm}

\usepackage{amsthm}
\usepackage{enumitem}

\usepackage{color}
\usepackage{soul}
\usepackage{multirow}
\usepackage{xcolor}
\usepackage{colortbl}
\usepackage{graphicx}
\graphicspath{{Source/}}

\definecolor{MyOrange}{HTML}{FF7F0E}
\definecolor{MyBlue}{HTML}{1F77B4}
\definecolor{lightRed}{HTML}{FD6C6C}
\definecolor{lightGreen}{HTML}{8AD187}

\newcommand{\etal}{\textit{et al.}}

\AtBeginDocument{%
  }

\copyrightyear{2024}
\acmYear{2024}
\setcopyright{rightsretained}
\acmConference[ICVGIP 2024]{Indian Conference on Computer Vision Graphics and Image Processing}{December 13--15, 2024}{Bengaluru, India}
\acmBooktitle{Indian Conference on Computer Vision Graphics and Image Processing (ICVGIP 2024), December 13--15, 2024, Bengaluru, India}
\acmDOI{10.1145/3702250.3702254}
\acmISBN{979-8-4007-1075-9/24/12}

\ccsdesc[500]{Information systems~Digital libraries and archives}

\begin{document}

\title{Lost in Context: The Influence of Context on Feature Attribution Methods for Object Recognition}

\author{Sayanta Adhikari}
\authornote{Both authors contributed equally to this research.}
\email{ai22mtech12005@iith.ac.in}
\orcid{0009-0008-3717-9223}
\affiliation{%
  \institution{Indian Institute of Technology Hyderabad}
  \city{Hyderabad}
  \state{Telangana}
  \country{India}
}

\author{Rishav Kumar}
\orcid{0009-0006-8335-4577}
\email{ai22mtech12003@iith.ac.in}
\authornotemark[1]
\affiliation{%
  \institution{Indian Institute of Technology Hyderabad}
  \city{Hyderabad}
  \state{Telangana}
  \country{India}
}

\author{Konda Reddy Mopuri}
\orcid{0000-0001-8894-7212}
\email{krmopuri@ai.iith.ac.in}
\affiliation{%
  \institution{Indian Institute of Technology Hyderabad}
  \city{Hyderabad}
  \state{Telangana}
  \country{India}
}

\author{Rajalakshmi Pachamuthu}
\orcid{0000-0002-7252-6728}
\email{raji@ee.iith.ac.in}
\affiliation{%
  \institution{TiHAN, Indian Institute of Technology Hyderabad}
  \city{Hyderabad}
  \state{Telangana}
  \country{India}
}

\renewcommand{\shortauthors}{Adhikari et al.}

\begin{abstract}
Contextual information plays a critical role in object recognition models within computer vision, where changes in context can significantly affect accuracy, underscoring models' dependence on contextual cues. This study investigates how context manipulation influences both model accuracy and feature attribution, providing insights into the reliance of object recognition models on contextual information as understood through the lens of feature attribution methods.


We employ a range of feature attribution techniques to decipher the reliance of deep neural networks on context in object recognition tasks. Using the ImageNet-9 and our curated ImageNet-CS datasets, we conduct experiments to evaluate the impact of contextual variations, analyzed through feature attribution methods. Our findings reveal several key insights: (a) Correctly classified images predominantly emphasize object volume attribution over context volume attribution. (b) The dependence on context remains relatively stable across different context modifications, irrespective of classification accuracy. (c) Context change exerts a more pronounced effect on model performance than Context perturbations. (d) Surprisingly, context attribution in `no-information' scenarios is non-trivial. Our research moves beyond traditional methods by assessing the implications of broad-level modifications on object recognition, either in the object or its context. Code available at \href{https://github.com/nineRishav/Lost-In-Context}{https://github.com/nineRishav/Lost-In-Context}
\end{abstract}

\ccsdesc[500]{Computing methodologies~Object detection}


\keywords{Context, Explainable AI (XAI), ImageNet, Feature Attribution, Object Recognition}



\maketitle

\section{Introduction}
\label{sec:intro}

In Computer Vision, \textit{Context} refers to information or signals not part of an object's appearance, including visual scenes and other objects. It encompasses various types of information, such as the presence and relationships of co-occurring objects, the overall scene type, ambient lighting conditions, and spatial arrangements, all of which contribute to a comprehensive understanding of the visual scene. Context plays a vital role in humans' ability to comprehend scene information. In situations where something is unclear, humans rely on contextual cues to understand the scene, such as when driving in foggy or rainy conditions, where visibility is low, but contextual information like the road layout and surrounding vehicles helps navigate safely.

\begin{figure}[h]
\centering
\includegraphics[width=0.85\linewidth]{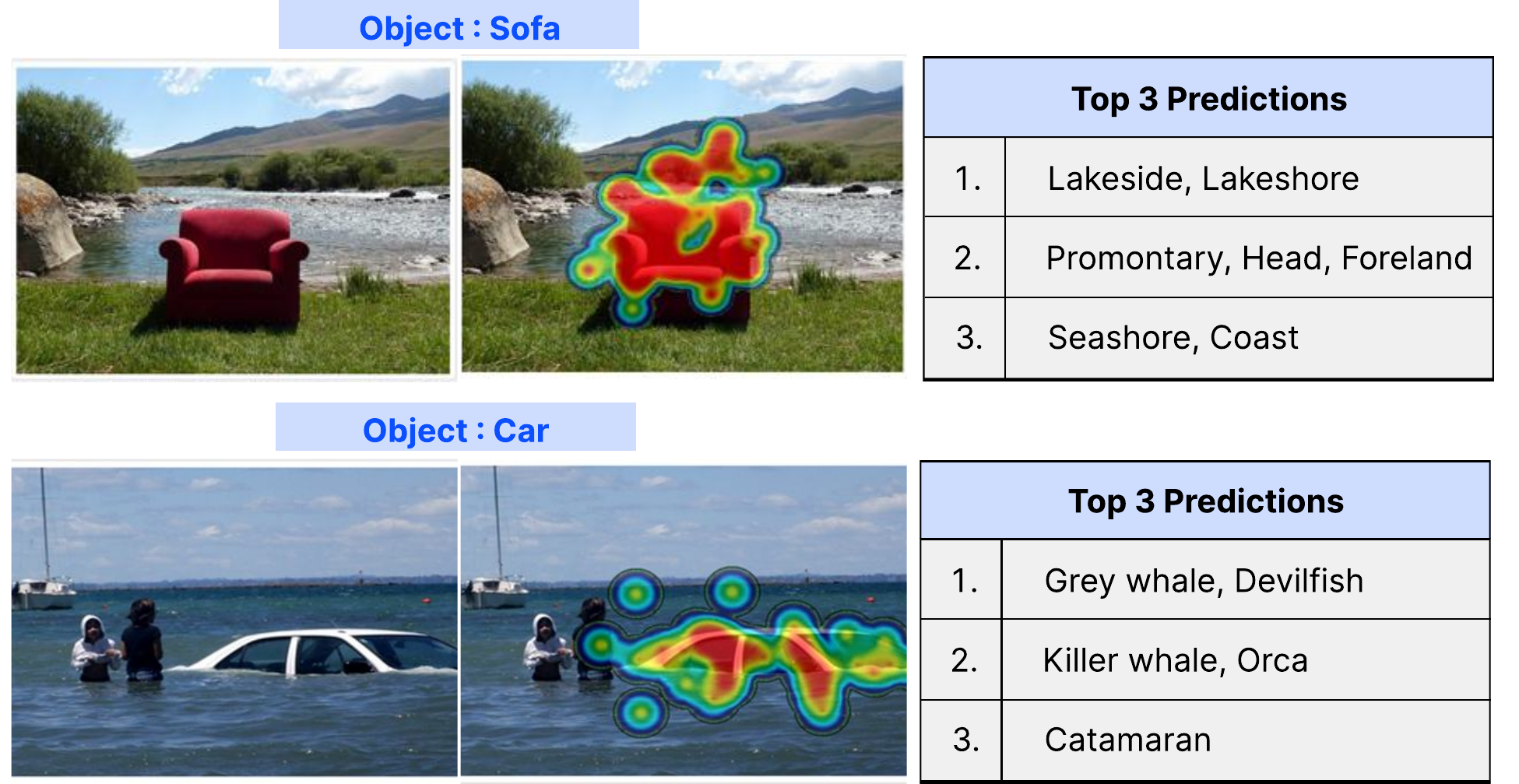}
\Description{An image showing the predictions emphasizing contextual information with feature attribution highlighting the object in the image.}
\caption{The predictions emphasize contextual information, while feature attribution for the top predicted class highlights the object in the image. Predictions were generated using a pre-trained \textbf{ResNet50} model, and feature attribution was performed with \textbf{GradCAM}.}
\label{fig:main_fig1}
\vspace{-10pt}
\end{figure}

The visual context of an object is an important source of information for recognition tasks in both human and computer vision. Understanding the importance of context in computer vision has become a significant field of study. One of the primary tasks in computer vision is Object Recognition, where deep learning models tend to surpass human-level performance \cite{dnn_passing_human}, making it an ideal area to study the effects of context as perceived by these models.

With the introduction of Deep Neural Networks (DNNs) in computer vision, their utility has grown exponentially. DNN models have surpassed traditional methods in various applications, including autonomous vehicles \cite{autonomous-driving, autonomous-benefit-1}, medical imaging \cite{medical, medical-survey}, face recognition \cite{face-recognition-benefit-1}, and motion detection \cite{motion-benefit-1}. In most cases, DNN models are used as black box models \cite{black-box}, with users primarily concerned about their performance. However, understanding how or why a model provides certain outputs is crucial for critical applications, such as medical imaging \cite{samek2017explainable}. For example, in Figure \ref{fig:main_fig1} (top row), feature attribution focuses on the object, a sofa, but the model's prediction is influenced by the context, i.e., lakeside, coast. These concerns have led to the introduction of explainability in machine learning and deep learning models. Many works \cite{paper2-mit+google, srinivas2019full, paper-4-google_brain, xai_debugging} have been conducted in the visual explanation domain for computer vision applications. These visual explanation methods provide the regions the model looks at to generate a particular output prediction.

Some past works show the effect of context on the performance of models. For instance, Xiao \etal\ \cite{xiao2021noise} have shown how context, or image background, plays a major role in the performance of models. Many past works \cite{benefit-context-1, benefit-context-2, benefit-context-3, benefit-context-4, benefit-context-5} have shown how context helps provide information in different ways to improve the model's accuracy.

This paper provides insights into the effect of contextual information on the model's prediction from the perspective of feature attribution methods. We extend this view by proposing methods and metrics to understand context effects using explainability methods. Explainability methods allow us to identify important areas, making it natural to use these methods to understand better how models perceive contextual information \cite{medical-survey}.

We have tried to interpret context effects such as context change and context perturbation from an explainability point of view and provide deeper insights using ImageNet-9 and ImageNet-CS (Sec \ref{sec:Im-cs-dataset}), where perturbation is added to the context. In this paper, we propose a simple metric to compute the importance of context as provided by popular feature attribution methods in the literature.

The major contributions of this paper are: 
\begin{itemize}
\vspace{-5pt}
\item We propose a novel approach to understand the effect of context on object recognition models through feature attribution methods. By leveraging feature attribution maps and segmentation maps, we introduce a new metric for quantitatively analyzing the impact of context (Section \ref{sec: metric}).

\item Our experimental results demonstrate that context change significantly affects model performance more than context perturbation (Section \ref{subsec:context-change-perturbation}).  Additionally, we show that models trained on larger datasets exhibit reduced reliance on contextual pixels (Section \ref{subsec:dataset}).

\item Our study on object size reveals that contrary to expectations, context attribution does not vary significantly between larger and smaller objects, indicating that object size alone does not predict context dependence. This challenges the assumption that more context inherently increases context reliance, highlighting the complexity of these interactions (Section \ref{subsec:object-size}).

\item We reveal that misclassification is closely linked to context change or perturbation. Furthermore, our experiments indicate that `no information' contexts tend to attract higher attribution than expected, highlighting the nuances of context influence (Section \ref{subsec:no-information}).


\end{itemize}

\section{Preliminaries}

\subsection{Models}

In Computer Vision, convolutional neural networks (CNNs) are fundamental, serving as the foundational components in prominent DNN models across diverse applications. The literature has witnessed significant advancements, leading to the introduction of various architectural designs that facilitate the extraction of discriminative features for accurate classification. Among these architectures, InceptionNet \cite{inception-net}, VGG \cite{vgg}, ResNet \cite{resnet}, and EfficientNet \cite{efficientnet} are some of the most notable.

ResNet (Residual Network) \cite{resnet} stands out due to its ability to accommodate deeper and larger structures by incorporating skip connections, which help mitigate the vanishing gradient problem and enable the training of very deep networks. In recent years, the introduction of transformer-based architectures in natural language processing (NLP) \cite{transformers} has also gained significant attention. Vision Transformer (ViT) \cite{vit}, a transformer-based architecture, has been widely adopted in computer vision tasks.

To explore various ideas and provide a holistic view of our study, we consider different architecture-based models, such as ResNet-50, ResNet-101, EfficientNet, and Vision Transformer (ViT-Base).\footnote{Pretrained weights for ResNet, EfficientNet, and ViT are taken from checkpoints provided by the \textit{timm} library \cite{timm}.}

\subsection{Feature Attribution Methods}
\label{sec: feat_attri}

\begin{figure*}[t]
    \centering
    \includegraphics[width=0.75\linewidth]{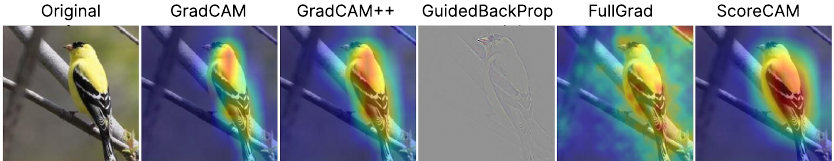}
    \caption{Comparison of feature attribution methods derived from ResNet50 classifier for top-1 prediction.}
    \label{fig:feature-attribution}
    \Description{A comparison of feature attribution methods derived from ResNet50 classifier for top-1 prediction.}
    \vspace{-10pt}
\end{figure*}

Understanding the perceptual cues that DNN models rely on for classification has consistently attracted the research community's interest. Within computer vision, the demand for visual explanations, particularly in identifying which regions of an image activate the network more, has led to the development of various visual explainability methods. The primary concept employed for interpreting models in this context is utilizing \textit{gradients}, which play a pivotal role in generating visual explanations. Notably, different explanation methods leverage gradients in diverse ways. We focus our study on these methods (See Figure \ref{fig:feature-attribution}). The visual explanation methods evaluated in this paper include:

\textbf{GradCAM} \cite{selvaraju2017grad} generalizes Class Activation Maps (CAM), which was initially applicable only to models with global average pooling as the penultimate layer. GradCAM aggregates the final convolutional layers' activation maps using the average gradients of the score concerning the activation maps, making it one of the most commonly used methods for visual explanations.

\textbf{GradCAM++} \cite{chattopadhay2018grad} improves GradCAM for better object localization by considering a weighted average of the gradients to find weights associated with each activation map of the penultimate layer. This method calculates the weights using higher-order score derivatives concerning activation maps.

\textbf{Guided Backpropagation} \cite{springenberg2014striving} combines backpropagation and deconvolution to compute the gradient of the score produced by the model for a particular class concerning the input image. It retains only the positive values of the output gradient, setting negative values to zero.

\textbf{FullGrad} \cite{srinivas2019full} emphasizes the importance of gradients across all layers of the architecture. Instead of visualizing a single layer, FullGrad aggregates gradients from all layers to provide comprehensive visual explanations.

\textbf{ScoreCAM} \cite{wang2020score} is a gradient-free approach that aggregates activation maps for visual explanations. Instead of using gradient sums as weights, ScoreCAM employs the class prediction score. Each activation map generates a masked input image, passed through the network to obtain the score for the observed class. These scores serve as weights for the activation maps, eliminating the dependency on gradients for visual explanations.

The selection criteria for the feature attribution methods in this study focused on their relevance and effectiveness in understanding vision models, particularly for object recognition. Guided Backpropagation uses gradients to provide fine-grained insights into input contributions. GradCAM and GradCAM++ offer more localized visual explanations by accumulating gradient-based activation maps, while FullGrad aggregates gradients across layers for comprehensive insights. In contrast, ScoreCAM uses score-based activation map accumulation, eliminating the need for gradients and providing a distinct perspective. Together, these methods offer diverse visualizations that enhance our understanding of feature attribution.


In contrast, methods like LIME \cite{LIME} and SHAP \cite{SHAP}, while effective for tabular data, lack smooth feature attribution capabilities for image analysis. Similarly, newer XAI methods like Concept Bottleneck Models (CBMs) \cite{CBM} focus on higher-level concepts rather than pixel-level attribution. While CBMs are valuable for understanding the influence of abstract concepts on model predictions, they do not align with our focus on fine-grained feature attribution in object recognition tasks. Hence, we excluded LIME, SHAP, and CBMs from consideration for this analysis to maintain a consistent focus on methods that provide detailed, spatially coherent visual explanations suitable for image data.

\section{Methodology}
\label{sec: methodology}

The primary aim of this work is to understand the effect of context on object recognition. To achieve this, we considered several pre-trained models: ResNet50, ResNet50-IN9L (trained on ImageNet-9), ResNet101, EfficientNet (pre-trained on ImageNet-21k and fine-tuned on ImageNet-1k) and ViT. We employed various feature attribution methods, including GradCAM, GradCAM++, ScoreCAM, Guided Backpropagation, and FullGrad. 


We used two datasets for our study: ImageNet-9 for context change and ImageNet-CS for context perturbation, as described in Section \ref{sec: datasets}. These datasets contain different context corruptions, providing a comprehensive view of the context's effect on object recognition. They also include segmentation maps for quantifying feature attribution on object and context.

To quantify feature attribution independently for objects and context, we proposed a metric detailed in Section \ref{sec: metric}. We first generated feature attribution maps for the images based on the top prediction given by each pre-trained model. We then calculated object and context volume attributions using the feature attribution maps and the segmentation map of each image. This process was repeated for different subsets of the dataset, and the averages were taken to provide holistic statistics for the dataset.


\section{Experimental Setup}
\subsection{Datasets}
\label{sec: datasets}

We utilized two datasets (see Figure \ref{fig:dataset}), applying perturbations and changes to the background while keeping the objects intact. The methodology involved measuring performance using predefined metrics and feature attribution methods. These two different datasets are manipulated versions of subsets of ImageNet-1k \cite{imagenet-1k}.

\subsubsection{\textbf{ImageNet-9} \cite{xiao2021noise}} 
\label{sec:Im-9-dataset}

\begin{figure*}[htbp]
    \centering
    \includegraphics[width=0.8\linewidth]{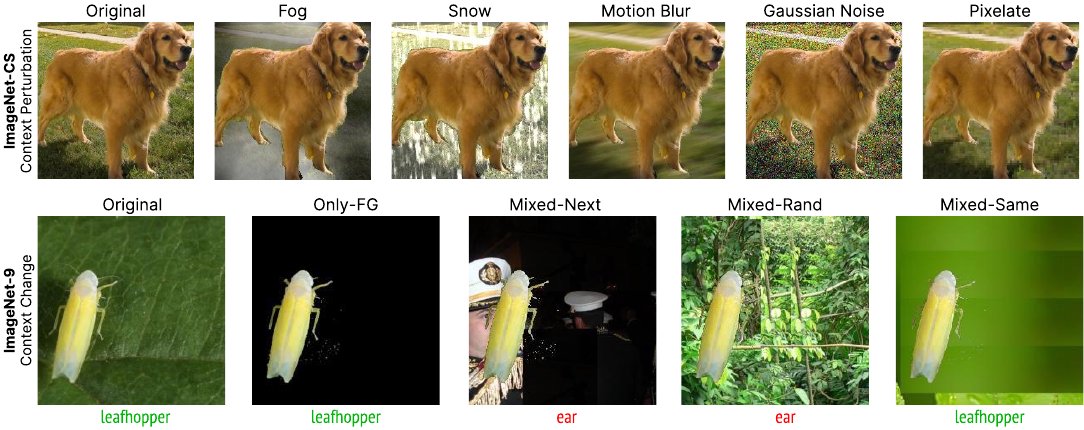}
    \Description{The top row provides images related to different varieties of our synthetic dataset named ImageNet-CS. The bottom row provides images showing variations of the ImageNet-9 dataset that we have considered for our experiments. We labeled ImageNet-9 images with its pre-trained ResNet50 classification-green, if corresponding with the original label; red, if not.}
    \caption{The top row provides images related to different varieties of our synthetic dataset named ImageNet-CS. The bottom row provides images showing variations of the ImageNet-9 dataset that we have considered for our experiments. We labelled ImageNet-9 images with its pre-trained ResNet50 classification—\textcolor{green!60}{green}, if corresponding with the original label; \textcolor{red!60}{red}, if not. }
    \label{fig:dataset}
    \vspace{-10pt}
\end{figure*}

The ImageNet-9 dataset comprises nine high-level classes covering 370 of the 1000 classes in the original ImageNet dataset. These high-level classes were generated using the WordNet \cite{wordnet} hierarchy. To study the effect of context, this dataset provides various manipulated contexts while keeping the object untouched. From this dataset, we have considered five different varieties of context changes (see the bottom row of Fig. \ref{fig:dataset}).

The dataset variations used for analysis are described as follows: \textbf{Original} contains the actual images from ImageNet-1k. \textbf{Only-FG} consists of only the object pixels, where the context pixels have been blacked out. In \textbf{Mixed-Next}, the context of the next class, based on the ImageNet-9 hierarchy, is added to the object. \textbf{Mixed-Rand} includes the context of a random class, which is added to the object. Lastly, \textbf{Mixed-Same} involves adding context from different images belonging to the same class to the object.

We examined these variations exclusively as they represent the entirety of available options, each altering the context in different ways. Our selection of classes for ImageNet-9 was based on the proportion of contextual pixels present in the images. Specifically, we focused on images where contextual pixels accounted for over 30\% of the total pixel count. Consequently, we included 89\% of the images in our analysis, resulting in a total of 3602 images. This dataset serves as a valuable resource for investigating how contextual factors influence object recognition across diverse environmental settings.

\subsubsection{\textbf{ImageNet-CS}}
\label{sec:Im-cs-dataset}

While ImageNet-9 provides images with changed contexts, it lacks variations where the context remains the same but is perturbed to some level. Additionally, ImageNet-9 only covers 37\% of the total ImageNet-1k classes, which might not provide generalized insights for our experiment. To address these gaps, we created our dataset, \textit{ImageNet-CS} (see the top row of Figure \ref{fig:dataset}), using two existing datasets: ImageNet-C (corruption) \cite{imagenet-c} and ImageNet-S (segmentation) \cite{imagenet-s}.

The ImageNet-C dataset provides various ways to corrupt the ImageNet classes, including categories like noise, blur, weather changes, and digital distortions, allowing us to apply different levels of corruption to the images. The ImageNet-S dataset provides segmentation maps for 919 classes of the ImageNet-1k dataset. By combining both datasets, we introduced ImageNet-CS, which allows for the independent manipulation of context and object. More details regarding this dataset are presented in the Supplementary document.
 
Our study sought to investigate image perturbations in the context of autonomous vehicles, focusing on conditions encountered during routine operations. To this end, we selected five types of corruption for analysis: \textit{fog, snow, motion blur, Gaussian noise, and pixelate}, representing a range of adverse weather conditions and camera malfunctions. To assess the effect of context, we included only images with contextual pixels comprising more than 30\% of the total pixels. Applying this constraint, we included 81.72\% of the total images, resulting in a dataset of 10,149 images for our analysis.

\subsection{Volume attribution metric}
\label{sec: metric}

This metric calculates the amount of feature attribution volume placed over the context and object. We require a feature attribution map and object segmentation mask to calculate the metric.
As feature attribution is not normalized, and different images may get different total feature attribution values, it is ideal to have a normalized metric. We normalized context volume attribution and object volume attribution by the total volume attribution of the image.

For an Image $\mathcal{I}$, we assume $\mathcal{A}$ to be the feature attribution map, and $\mathcal{M}_{\mathcal{I}}$ as the binary object segmentation mask. We denote $V_O^{\mathcal{I}}$ as volume attribution for the Object and $V_C^{\mathcal{I}}$ as volume attribution for the Context of a particular image $\mathcal{I}$.

\begin{equation}
    V_O^{\mathcal{I}} = \frac{\sum (\mathcal{A} \odot \mathcal{M}_{\mathcal{I}})}{\sum \mathcal{A}} \;\; , \;\;\;
    V_C^{\mathcal{I}} = \frac{\sum (\mathcal{A} \odot (1 - \mathcal{M}_{\mathcal{I}}))}{\sum \mathcal{A}}
    \label{eq:vol_context}
\end{equation}

In equation \ref{eq:vol_context}, $\odot$ denotes Hadamard Product. This helps us to separate object and context attributions for an image. To study the general statistics over a dataset, we can take the average of all the images of that dataset to get average context and object volume attribution. This metric is easy to compute and can indicate quantitatively how important context and object are to a particular model and a feature attribution method.

\section{Observations}
\label{sec: observations}

\begin{table*}[htbp]
\centering


\caption{Accuracy (in percentage) of ResNet50, ResNet50-IN9L, ResNet101, EfficientNet, and Vision Transformer (ViT) on ImageNet-9 and ImageNet-CS samples. Original accuracy (Orig.) is shown, along with accuracy under context changes (only-fg, mixed-next, mixed-rand, mixed-same) and perturbations (fog, snow, motion blur, gaussian noise, pixelate). ImageNet-CS perturbations were applied to ImageNet-9 samples. The lowest and highest values are highlighted in \textcolor{lightRed}{red} and \textcolor{lightGreen}{green}. Mean CC and Mean CP represent average accuracies for context changes (CC) and perturbations (CP), respectively.}

\vspace{-5pt}

\begin{tabular}{@{}l|c|crrr|crrrr|rr@{}}
\toprule
\multicolumn{1}{c|}{\multirow{2}{*}{\textbf{Model}}} &
  \multirow{2}{*}{\textbf{Orig.}} &
  \multicolumn{4}{c|}{\cellcolor{cyan!20}\textbf{Context Change (CC)}} &
  \multicolumn{5}{c|}{\cellcolor{yellow!20}\textbf{Context Perturbation (CP)}} &
  \multicolumn{2}{c}{\textbf{Avg. Decline from Orig.}} \\ \cmidrule(lr){3-13} 
  \multicolumn{1}{c|}{} & & only-fg &
  \multicolumn{1}{c}{\begin{tabular}[c]{@{}c@{}}mixed-\\ next\end{tabular}} &
  \multicolumn{1}{c}{\begin{tabular}[c]{@{}c@{}}mixed-\\ rand\end{tabular}} &
  \multicolumn{1}{c|}{\begin{tabular}[c]{@{}c@{}}mixed-\\ same\end{tabular}} &
  fog &
  \multicolumn{1}{c}{snow} &
  \multicolumn{1}{c}{\begin{tabular}[c]{@{}c@{}}motion \\ blur\end{tabular}} &
  \multicolumn{1}{c}{\begin{tabular}[c]{@{}c@{}}gaussian \\ noise\end{tabular}} &
  \multicolumn{1}{c|}{pixelate} &
  \multicolumn{1}{c}{\cellcolor{cyan!20}\begin{tabular}[c]{@{}c@{}}Orig. -\\Mean CC\end{tabular}} &
  \multicolumn{1}{c}{\cellcolor{yellow!20}\begin{tabular}[c]{@{}c@{}}Orig. - \\Mean CP\end{tabular}} \\ \midrule
\textbf{ResNet50} &
  \multicolumn{1}{c|}{95.9} &
  \multicolumn{1}{c}{88.1} &
  \multicolumn{1}{c}{\textcolor{lightRed}{82.1}} &
  \multicolumn{1}{c}{83.8} &
  \multicolumn{1}{c|}{\textcolor{lightGreen}{89.6}} &
  \multicolumn{1}{c}{93.4} &
  \multicolumn{1}{c}{92.5} &
  \multicolumn{1}{c}{93.6} &
  \multicolumn{1}{c}{93.3} &
  \multicolumn{1}{c|}{\textcolor{lightGreen}{94.1}} &
  \multicolumn{1}{c}{10.0} &
  \multicolumn{1}{c}{2.5} \\
\textbf{ResNet50-IN9L} &
  \multicolumn{1}{c|}{95.8} &
  \multicolumn{1}{c}{84.5} &
  \multicolumn{1}{c}{\textcolor{lightRed}{72.2}} &
  \multicolumn{1}{c}{76.2} &
  \multicolumn{1}{c|}{\textcolor{lightGreen}{90.7}} &
  \multicolumn{1}{c}{92.9} &
  \multicolumn{1}{c}{89.8} &
  \multicolumn{1}{c}{92.4} &
  \multicolumn{1}{c}{83.0} &
  \multicolumn{1}{c|}{\textcolor{lightGreen}{93.8}} &
  \multicolumn{1}{c}{14.9} &
  \multicolumn{1}{c}{5.4} \\ 
\textbf{ResNet101} &
  \multicolumn{1}{c|}{96.9} &
  \multicolumn{1}{c}{91.0} &
  \multicolumn{1}{c}{\textcolor{lightRed}{84.8}} &
  \multicolumn{1}{c}{86.1} &
  \multicolumn{1}{c|}{\textcolor{lightGreen}{91.4}} &
  \multicolumn{1}{c}{94.3} &
  \multicolumn{1}{c}{94.1} &
  \multicolumn{1}{c}{94.7} &
  \multicolumn{1}{c}{94.4} &
  \multicolumn{1}{c|}{\textcolor{lightGreen}{95.5}} &
  \multicolumn{1}{c}{8.6} &
  \multicolumn{1}{c}{2.3} \\ 
\textbf{EfficientNet} &
  \multicolumn{1}{c|}{95.4} &
  \multicolumn{1}{c}{84.4} &
  \multicolumn{1}{c}{\textcolor{lightRed}{74.1}} &
  \multicolumn{1}{c}{76.8} &
  \multicolumn{1}{c|}{\textcolor{lightGreen}{86.2}} &
  \multicolumn{1}{c}{91.9} &
  \multicolumn{1}{c}{91.1} &
  \multicolumn{1}{c}{92.4} &
  \multicolumn{1}{c}{89.6} &
  \multicolumn{1}{c|}{\textcolor{lightGreen}{93.7}} &
  \multicolumn{1}{c}{15.0} &
  \multicolumn{1}{c}{3.7} \\ 
\textbf{ViT (base)} &
  \multicolumn{1}{c|}{96.5} &
  \multicolumn{1}{c}{84.9} &
  \multicolumn{1}{c}{\textcolor{lightRed}{82.5}} &
  \multicolumn{1}{c}{83.7} &
  \multicolumn{1}{c|}{\textcolor{lightGreen}{89.3}} &
  \multicolumn{1}{c}{95.1} &
  \multicolumn{1}{c}{91.3} &
  \multicolumn{1}{c}{94.8} &
  \multicolumn{1}{c}{94.2} &
  \multicolumn{1}{c|}{\textcolor{lightGreen}{95.9}} &
  \multicolumn{1}{c}{11.4} &
  \multicolumn{1}{c}{2.2} \\ \bottomrule
\end{tabular}
\label{table: accuracy}
\vspace{-10pt}
\end{table*}

\subsection{The Impact of Context Change vs. Context Perturbations on Model Performance}
\label{subsec:context-change-perturbation}

As outlined in Section \ref{sec: datasets}, we categorize context corruptions into two types: \textbf{context change} and \textbf{context perturbation}. We applied these corruptions uniformly across all samples to assess their impact on the performance of various pre-trained models. The results, detailed in Table \ref{table: accuracy}, reveal significant differences in how models handle these two corruptions.

The experiments demonstrate that complete context changes result in more substantial performance declines than context perturbations. This finding indicates a strong reliance of models on contextual information for accurate object recognition. For instance, ResNet50 experiences an average performance decline of 10\% with context changes, which is notably higher than the 2.5\% decline observed with context perturbations.

In the ImageNet-9 dataset experiments (see Fig. \ref{fig: Qualitative Comparison}b), the \textit{`mixed-next'} variety causes significant accuracy drops due to poor contextual congruence; the background from an adjacent class introduces semantic confusion, impacting model performance. Conversely, \textit{`mixed-same'} and \textit{`pixelate'} maintain higher accuracy as they preserve contextual congruence. \textit{`mixed-same'} uses related backgrounds within the same category, aiding accurate object recognition, while \textit{`pixelate'} reduces detail but retains structural integrity. This underscores the crucial role of contextual alignment in visual recognition tasks \cite{pixelate-explain}.

Moreover, the data shows that models trained on more comprehensive datasets exhibit enhanced generalizability and resilience to context changes. In contrast, models like ResNet50-IN9L, trained on more limited datasets, tend to rely heavily on contextual shortcuts. This reliance limits their ability to generalize and exposes them to frequency shortcuts \cite{frequency-shortcuts}, where models capitalize on predominant training patterns at the expense of broader applicability. These findings underscore the critical need for robust models capable of handling diverse and changing contexts, particularly in applications where accurate object recognition is vital, such as autonomous driving and medical imaging. This stresses the importance of adopting training strategies prioritising intrinsic object features over contextual familiarity, ensuring models perform reliably in varied real-world scenarios.

\subsection{Impact of the Training Dataset Size}
\label{subsec:dataset}

Building on the observations from Section \ref{subsec:context-change-perturbation}, this section explores how the size of the training dataset influences model sensitivity to contextual cues. Models trained on extensive datasets, such as ResNet50, typically demonstrate a lower dependency on contextual information for accurate object recognition than those trained on limited datasets like ResNet50-IN9L.

\begin{figure}[ht]
    \includegraphics[width=0.90\linewidth]{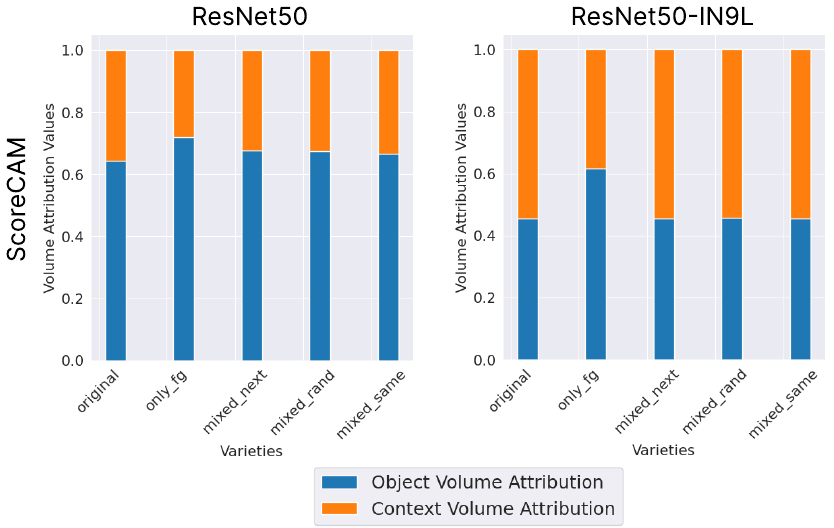}
    \Description{Comparison of object and context volume attribution for ResNet50 and ResNet50-IN9L on the ImageNet-9 dataset using ScoreCAM.}
    \caption{Comparison of \textcolor{MyBlue!80}{object} and \textcolor{MyOrange!80}{context} volume attribution for ResNet50 and ResNet50-IN9L on the ImageNet-9 dataset using ScoreCAM. The figure shows higher context reliance in ResNet50-IN9L, highlighting the impact of training dataset size on model sensitivity to contextual cues.}
    \label{fig:vol_attri}
    \vspace{-10pt}
\end{figure}

As illustrated in Fig. \ref{fig:vol_attri} and detailed in Table \ref{table: accuracy}, there is a noticeable distinction in context attribution between these models. Specifically, the average context attribution for ResNet50-IN9L exceeds 60\% across all tested scenarios, significantly higher than ResNet50, which remains below 40\%. This quantifies how dataset breadth impacts model reliance on context for decision-making.

To validate these observations, we employed various visualization methods, including ScoreCAM, to analyze model behaviour across different pre-trained models. This comprehensive approach confirms that models with broader training data rely less on contextual cues, enhancing their generalizability.

Additional feature attribution plots, which provide deeper insights into model decision processes, are available in the Supplementary document. These visualizations further corroborate our findings, underscoring the critical role of training data volume in developing robust machine-learning models.

\subsection{Context Influence on Misclassification}
\label{subsec:misclassification}

\begin{figure*}[ht]
    \centering
    \includegraphics[width=0.9\textwidth]{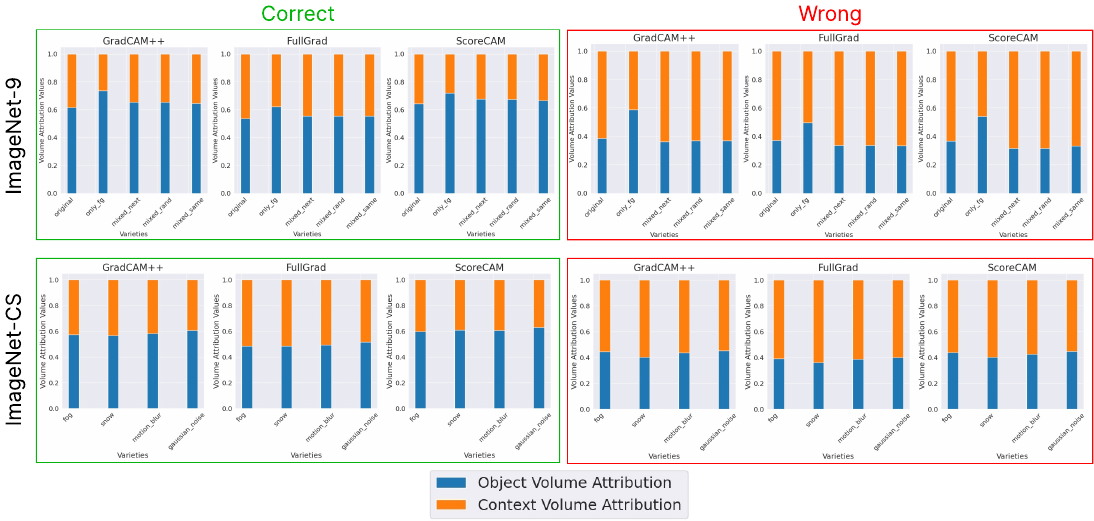}
    \Description{This plot illustrates the variation in volume attribution of Context Correctly Classified Set and Wrongly Classified Set classifications, using ResNet50 as our backbone architecture. The feature attributions were generated based on the model's top prediction. The volume attributions for both ImageNet-9 and ImageNet-CS datasets are presented.}
    \caption{This plot illustrates the variation in \textcolor{MyOrange!80}{volume attribution of Context} for \textcolor{green!60}{Correctly Classified Set} and \textcolor{red!60}{Wrongly Classified Set} classifications, using ResNet50 as our backbone architecture. The feature attributions were generated based on the model's top prediction. The volume attributions for both ImageNet-9 and ImageNet-CS datasets are presented.}
    \label{fig: misclassification}
    \vspace{-10pt}
\end{figure*}

Misclassification in machine learning models often correlates strongly with changes in context, as detailed in Section \ref{subsec:context-change-perturbation}. Higher context volume attributions are typically observed in misclassified instances than those with more prominent object volume. To explore this correlation further, we partitioned our datasets into two sets based on the classification outcomes of the pre-trained models: The \textbf{Correctly Classified Set} comprises images accurately classified by the model across different ImageNet-9 dataset varieties. In contrast, the \textbf{Wrongly Classified Set} includes images misclassified by the pre-trained model within each variety.

This segregation serves the dual purpose of examining the feature attribution method's ability to differentiate between contexts when the model achieves consistent accuracy across varieties and assessing the impact of context on misclassification in the Wrongly Classified Set. Figure \ref{fig: misclassification} provides insights into these aspects.

The analysis reveals that in the Wrongly Classified Set, context volume attributions for the ImageNet-9 and ImageNet-CS variants are approximately 20\% and 10\% higher, respectively, than those in the Correctly Classified Set. This significant difference underscores a high correlation between increased context attributions and the likelihood of misclassification, suggesting that context perturbations substantially impact model predictions.

Furthermore, the examination of the Correctly Classified Set indicates that object-context volume attributions remain consistent across contexts, highlighting that feature attribution methods struggle to distinguish irrelevant contexts (\textit{`original', `mixed\_next', `mixed\_rand', `mixed\_same'}) effectively. This consistency across varied contexts where contextual information should be less impactful suggests potential areas for improving the robustness and accuracy of feature attribution methodologies in machine learning models.

\subsection{Impact of Object Size on Contextual Attribution}
\label{subsec:object-size}

In our ongoing exploration, we differentiated between objects of varying sizes to probe potential differences in model behaviour. We categorized `larger objects' as samples where the object occupies 30\% to 50\% of the image space, with the remaining being context, and `smaller objects' where the object accounts for less than 20\% of the image, thus having 80\% or more context pixels. It was considered that smaller objects, with their higher proportion of context, might reveal different patterns of contextual attribution.

\begin{figure}
\centering
\includegraphics[width=0.76\linewidth]{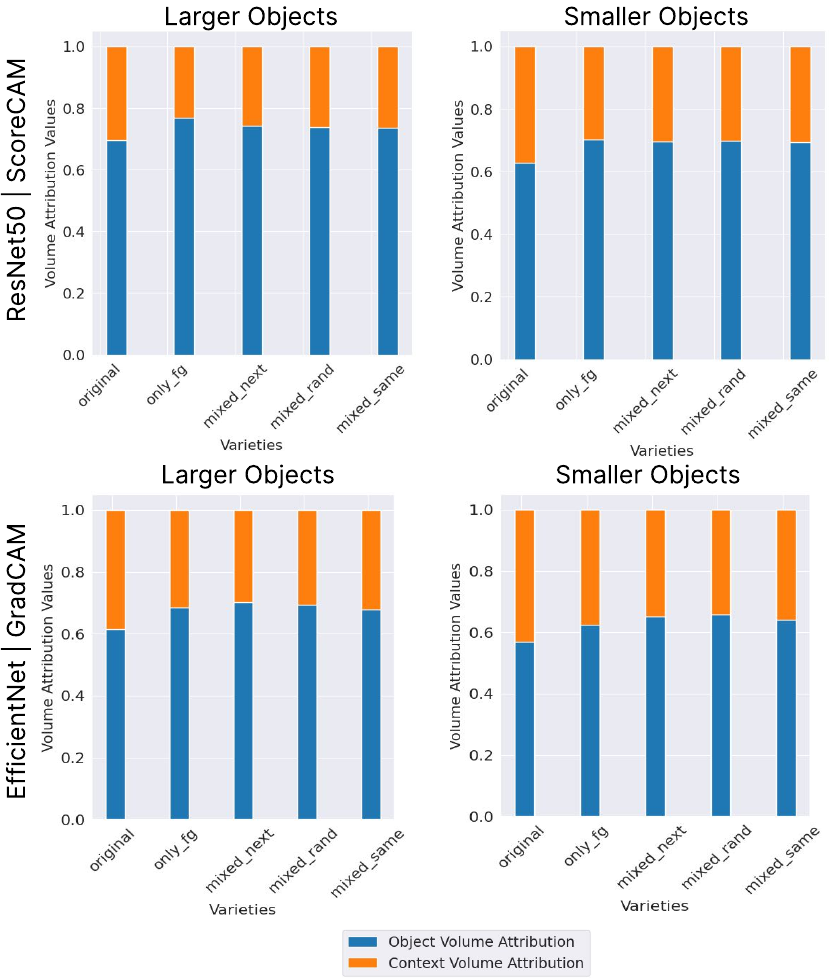}
\caption{Average volume attribution for larger vs. smaller objects in the ImageNet-9 dataset. The plot indicates that \textcolor{MyOrange!80}{context volume attribution} is similar across object sizes, suggesting that more context pixels do not necessarily increase context influence in model predictions.}
\Description{Comparison of average volume attribution between larger and smaller objects within the ImageNet-9 dataset. The plot shows that context volume attribution does not significantly differ between object sizes.}\label{fig: big_small}
\vspace{-15pt}
\end{figure}

However, as depicted in Fig. \ref{fig: big_small}, the data did not support our initial contemplation. The figure shows that object-context attribution for larger and smaller objects remains comparable, thus questioning the idea that more context pixels automatically lead to a greater influence of context in model predictions.

In conclusion, our exploration suggests that the size of an object does not have a marked impact on how a model utilizes context in its predictions. This nuanced finding points to the complexity of the relationship between object size, context, and model performance, suggesting that these dynamics warrant further investigation.

\begin{figure*}
    \centering
    \includegraphics[width=0.95\linewidth]{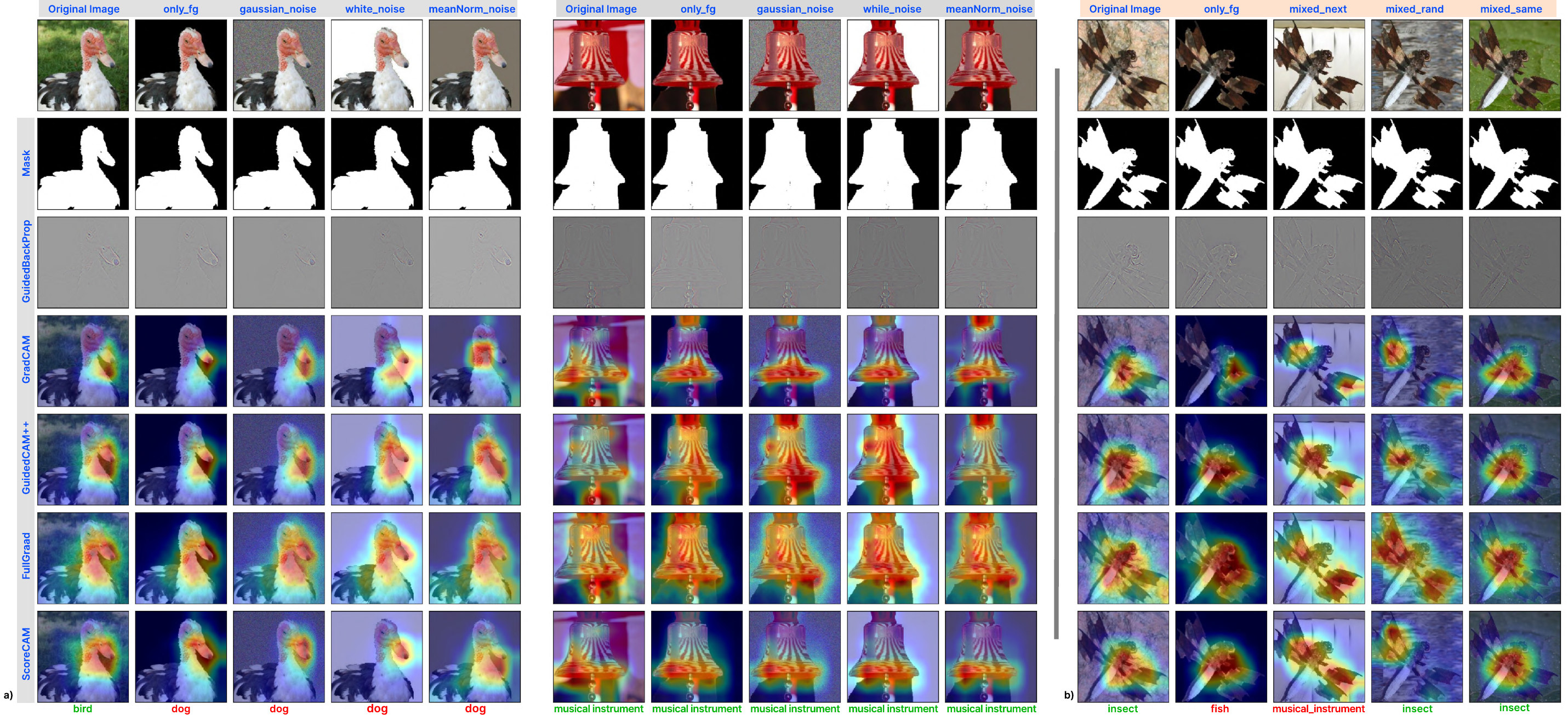}
    \caption{(a) Qualitative Feature Attribution Comparison of `No-Information' variants\newline
    \hspace*{4.15em}(b) Qualitative comparison of \textit{Original, only\_fg, mixed\_next, mixed\_rand, and mixed\_same variants.}}
    \Description{(a) Qualitative Feature Attribution Comparison of `No-Information' variants, (b) Qualitative comparison of Original, only-fg, mixed-next, mixed-rand, and mixed-same variants.}
    \label{fig: Qualitative Comparison}
\end{figure*}



\subsection{`No Information' Context Varieties}
\label{subsec:no-information}

Within the ImageNet-9 dataset, the \textit{only\_fg} category replaces all context with black pixels, ostensibly providing `no information' to the model. Preferably, this should result in minimal context attribution, ideally near zero. However, as demonstrated in Figures \ref{fig:vol_attri}, \ref{fig: misclassification}, and \ref{fig: big_small}, context attribution in these cases unexpectedly exceeds 30\% of \textit{only\_fg}.

\begin{figure}[ht]
    \vspace{-5pt}
    \centering
    \includegraphics[width=0.90\linewidth]{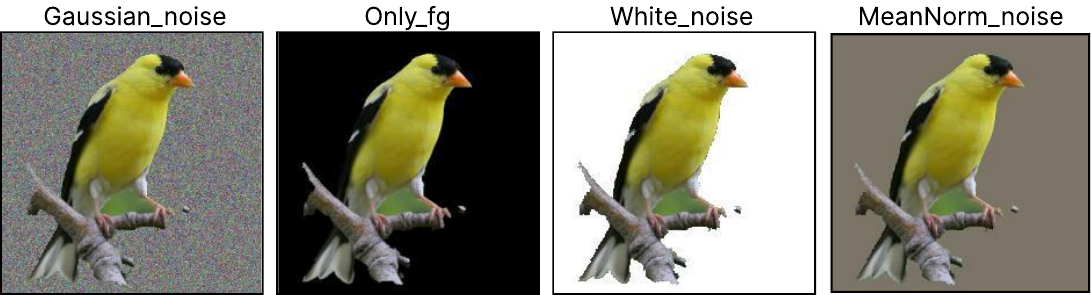}
    \Description{No Information context variants: Gaussian Noise, Only-fg, white-noise, meannorm-noise}
    \caption{No Information context variants: \textit{Gaussian\_noise}, \textit{Only\_fg}, \textit{White\_noise} and \textit{MeanNorm\_noise}}
    \label{fig: noise-data}
    \vspace{-10pt}
\end{figure}


\begin{figure}[ht]
    \centering
    \includegraphics[width=0.85\linewidth]{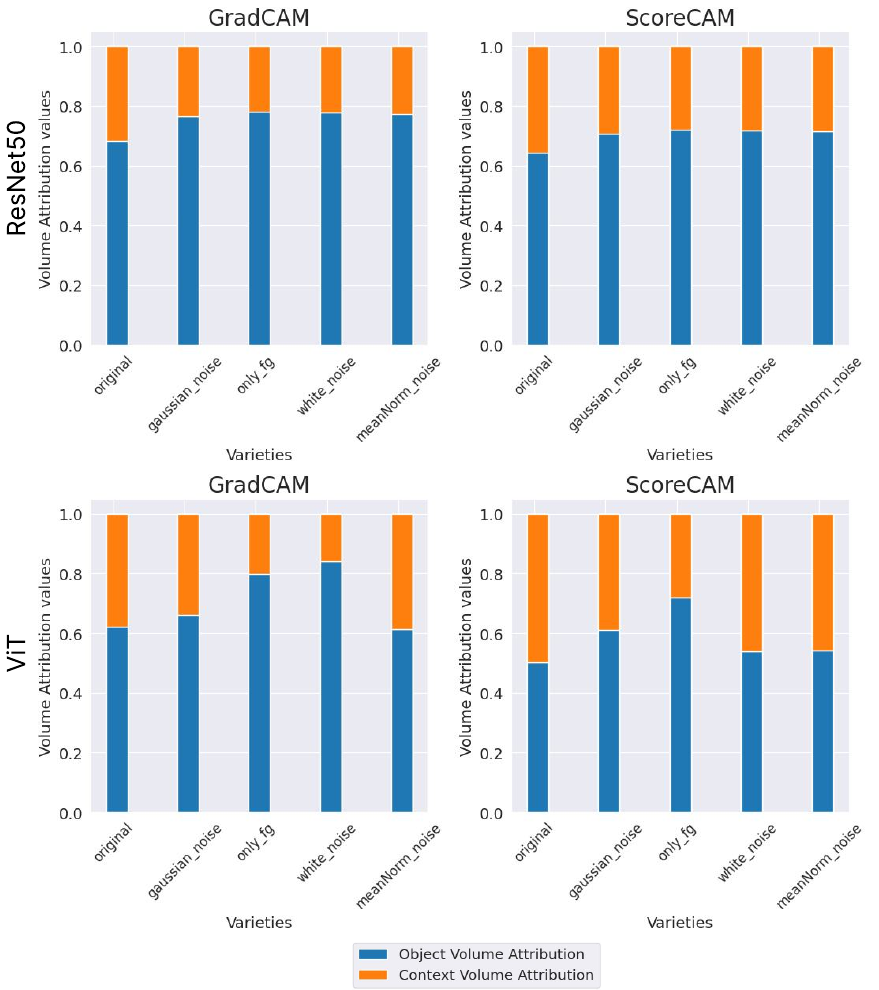}
    \caption{Feature attributions for `no information' context varieties. Note that there is a nontrivial \textcolor{orange!80}{context volume attribution}
 (30\%) despite having `no information' context from \textit{\text{only\_fg, white\_noise, meanNorm\_noise}}}
 \Description{Feature attributions for `no information' context varieties. Note that there is a nontrivial context volume attribution (30\%) despite having `no information' context from only_fg, white_noise, meanNorm_noise }
    \label{fig: noise-plot}
    \vspace{-15pt}
\end{figure}

To further investigate, we introduced variants (see Fig. \ref{fig: noise-data}) such as \textit{gaussian\_noise}, \textit{white\_noise}, and \textit{meanNorm\_noise}, where backgrounds are replaced with noise or mean values, offering similarly minimal informational content. Despite these changes, as visualized in Figures \ref{fig: Qualitative Comparison}a and \ref{fig: noise-plot}, context attribution remains consistently high across all `no information' variants. This indicates that the attribution process consistently assigns importance to context, irrespective of the actual context content.


This phenomenon was observed across both CNNs and transformer-based models, highlighting a fundamental issue in feature attribution methods, as shown in Figure \ref{fig: noise-plot}. Consistent with previous studies like Wilming \etal\ \cite{no-information-supporter-backup}, these methods often misattribute importance to \textit{suppressor variables}, which have no statistical link to the target prediction. This raises concerns about their reliability in distinguishing relevant from irrelevant information.

Building upon these concerns, recent insights from Adebayo \etal\ \cite{no-information-sanity-1} further underscore the need for more rigorous evaluations of how saliency maps and similar tools analyze the importance of features, especially in scenarios intentionally devoid of informative content. This is consistent with our observations where models frequently misinterpret `no information' contexts as significant, despite their lack of contributory value to decision-making processes. Similarly, the work by Ilyas \etal\ \cite{no-information-adversarial-2} provides a theoretical basis for our findings, suggesting that such misattributions may stem from models' reliance on features that are statistically useful but semantically meaningless. These insights necessitate reevaluating how feature importance is assessed, emphasizing the need to refine these methodologies to avoid misleading interpretations and improve the reliability of machine learning models in complex analytical tasks.

\section{Related Works}

The utilization of contextual information significantly enhances performance across various tasks in computer vision, including image recognition, segmentation, and scene understanding \cite{benefit-context-1, benefit-context-2, benefit-context-3, benefit-context-4}. Context helps distinguish visually similar objects, thereby improving object recognition accuracy, and plays a pivotal role in image segmentation by clarifying object boundaries in complex scenarios like occlusion \cite{tian2023learning}.

Despite these advantages, incorporating context is not universally beneficial and can sometimes lead to suboptimal model performance \cite{disadvantage-context-1, disadvantage-context-2}. A balanced approach that involves selective incorporation of context has been proposed to mitigate such issues \cite{context-selective}. This approach highlights the challenges in effectively leveraging context, particularly when models zoom into images, potentially overlooking valuable contextual cues \cite{zoom-required}.

Existing datasets like ObjectNet and ImageNet-9 are designed to test model resilience against background perturbations \cite{object-net-random_background, xiao2021noise}. However, there remains a gap in datasets that allow for controlled background changes without affecting the object, highlighting limitations in experimental setups that fail to separate the influences of object and background changes effectively \cite{feature-attribution-closest-to-us}.

Current research on feature attribution methods often focuses solely on the significance of input features, neglecting broader contextual changes. Studies have demonstrated that while foreground perturbations significantly impair model performance, background changes have a subtler effect, suggesting models retain effectiveness when trained with diverse backgrounds \cite{feature-attribution-closest-to-us, object_with_or_without_object}. This indicates potential robustness against certain types of contextual modifications but also highlights a gap in how models handle complex visual environments.

Insights from Adebayo \etal\ \cite{paper2-mit+google} and Yang \etal\ \cite{paper-4-google_brain} challenge the reliability of popular feature attribution tools like saliency maps, which often misrepresent the causal impact of features on model decisions, particularly in scenarios influenced by complex background information. Yang \etal\ further demonstrate that attribution methods vary in their ability to prioritize features accurately, often failing to align model interpretations with actual feature relevancy, especially under adversarial conditions or ambiguous contexts. These studies underline the need for more nuanced attribution methodologies that ensure AI systems remain interpretable and reliable across varied operational settings.

Our study extends beyond traditional approaches by examining the implications of coarse-level changes — either in the Object or Context—on model outputs and feature attribution methods. We critically analyze how these methods assess the importance of context and address potential misattribution to suppressor variables—features that do not contribute meaningfully to the model's decision-making process \cite{no-information-supporter, no-information-supporter-backup}. This work is vital as it informs the ongoing debate within the XAI community about ensuring model interpretability and reliability, particularly when distinguishing between relevant and irrelevant information is essential \cite{xai_debugging}.

\section{Conclusion}

This research highlights the crucial role of contextual information in object recognition, showing that context changes significantly impact model performance more than perturbations. Models trained on extensive datasets exhibited lower context dependency, emphasizing the importance of comprehensive training data for robustness. Furthermore, our study on feature attribution methods revealed a concerning trend: even in `no information' contexts, models often misattribute importance to irrelevant contextual pixels. These insights are vital for advancing explainable AI, particularly in critical applications like autonomous driving and medical imaging, advocating for training strategies that prioritize intrinsic object features and refining attribution methods to ensure robust model interpretability.

\section{Limitations \& Future Work}

Our study offers valuable insights but acknowledges certain constraints. Although the datasets used are comprehensive, they are subsets of the larger ImageNet dataset and might not encompass all real-world scenarios. Due to computational challenges, experimenting with a larger dataset is difficult. Moreover, our analysis primarily focused on ResNet, EfficientNet, and Vision Transformer models, and the results are based on metric-dependent feature attribution methods, which may not capture all nuances of context influence.

To advance this field, we propose integrating context change augmentations into training regimens to reduce models' reliance on irrelevant context cues and enhance their focus on salient object features. Future work should expand the range of datasets, explore other deep learning architectures, and develop more sophisticated attribution techniques. 


Overall, our research contributes crucial insights into the functionality and limitations of feature attribution methods, paving the way for more interpretable and reliable artificial intelligence systems.


\bibliography{acmbib}


\begin{thebibliography}{51}


\ifx \showCODEN    \undefined \def \showCODEN     #1{\unskip}     \fi
\ifx \showDOI      \undefined \def \showDOI       #1{#1}\fi
\ifx \showISBNx    \undefined \def \showISBNx     #1{\unskip}     \fi
\ifx \showISBNxiii \undefined \def \showISBNxiii  #1{\unskip}     \fi
\ifx \showISSN     \undefined \def \showISSN      #1{\unskip}     \fi
\ifx \showLCCN     \undefined \def \showLCCN      #1{\unskip}     \fi
\ifx \shownote     \undefined \def \shownote      #1{#1}          \fi
\ifx \showarticletitle \undefined \def \showarticletitle #1{#1}   \fi
\ifx \showURL      \undefined \def \showURL       {\relax}        \fi
\providecommand\bibfield[2]{#2}
\providecommand\bibinfo[2]{#2}
\providecommand\natexlab[1]{#1}
\providecommand\showeprint[2][]{arXiv:#2}

\bibitem[Adadi and Berrada(2018)]%
        {black-box}
\bibfield{author}{\bibinfo{person}{Amina Adadi} {and} \bibinfo{person}{Mohammed Berrada}.} \bibinfo{year}{2018}\natexlab{}.
\newblock \showarticletitle{Peeking Inside the Black-Box: A Survey on Explainable Artificial Intelligence (XAI)}.
\newblock \bibinfo{journal}{\emph{IEEE Access}}  \bibinfo{volume}{6} (\bibinfo{year}{2018}), \bibinfo{pages}{52138--52160}.
\newblock
\urldef\tempurl%
\url{https://doi.org/10.1109/ACCESS.2018.2870052}
\showDOI{\tempurl}


\bibitem[Adebayo et~al\mbox{.}(2018)]%
        {no-information-sanity-1}
\bibfield{author}{\bibinfo{person}{Julius Adebayo}, \bibinfo{person}{Justin Gilmer}, \bibinfo{person}{Michael Muelly}, \bibinfo{person}{Ian Goodfellow}, \bibinfo{person}{Moritz Hardt}, {and} \bibinfo{person}{Been Kim}.} \bibinfo{year}{2018}\natexlab{}.
\newblock \showarticletitle{Sanity Checks for Saliency Maps}.
\newblock \bibinfo{journal}{\emph{Advances in Neural Information Processing Systems (NeurIPS)}}  \bibinfo{volume}{31} (\bibinfo{year}{2018}).
\newblock


\bibitem[Adebayo et~al\mbox{.}(2020)]%
        {paper2-mit+google}
\bibfield{author}{\bibinfo{person}{Julius Adebayo}, \bibinfo{person}{Michael Muelly}, \bibinfo{person}{Ilaria Liccardi}, {and} \bibinfo{person}{Been Kim}.} \bibinfo{year}{2020}\natexlab{}.
\newblock \showarticletitle{Debugging Tests for Model Explanations}.
\newblock \bibinfo{journal}{\emph{arXiv preprint arXiv:2011.05429}} (\bibinfo{year}{2020}).
\newblock
\urldef\tempurl%
\url{https://arxiv.org/abs/2011.05429}
\showURL{%
\tempurl}
\newblock
\shownote{arXiv preprint}.


\bibitem[Alsallakh et~al\mbox{.}(2021)]%
        {xai_debugging}
\bibfield{author}{\bibinfo{person}{Bilal Alsallakh}, \bibinfo{person}{Narine Kokhlikyan}, \bibinfo{person}{Vivek Miglani}, \bibinfo{person}{Shubham Muttepawar}, \bibinfo{person}{Edward Wang}, \bibinfo{person}{Sara Zhang}, \bibinfo{person}{David Adkins}, {and} \bibinfo{person}{Orion Reblitz-Richardson}.} \bibinfo{year}{2021}\natexlab{}.
\newblock \showarticletitle{Debugging the Internals of Convolutional Networks}. In \bibinfo{booktitle}{\emph{eXplainable AI Approaches for Debugging and Diagnosis}}.
\newblock


\bibitem[Barbu et~al\mbox{.}(2019)]%
        {object-net-random_background}
\bibfield{author}{\bibinfo{person}{Andrei Barbu}, \bibinfo{person}{David Mayo}, \bibinfo{person}{Julian Alverio}, \bibinfo{person}{William Luo}, \bibinfo{person}{Christopher Wang}, \bibinfo{person}{Dan Gutfreund}, \bibinfo{person}{Josh Tenenbaum}, {and} \bibinfo{person}{Boris Katz}.} \bibinfo{year}{2019}\natexlab{}.
\newblock \showarticletitle{ObjectNet: A Large-Scale Bias-Controlled Dataset for Pushing the Limits of Object Recognition Models}. In \bibinfo{booktitle}{\emph{Advances in Neural Information Processing Systems (NeurIPS)}}, Vol.~\bibinfo{volume}{32}.
\newblock
\urldef\tempurl%
\url{https://proceedings.neurips.cc/paper/2019/file/97af07a14cacba681feacf3012730892-Paper.pdf}
\showURL{%
\tempurl}


\bibitem[Chattopadhay et~al\mbox{.}(2018)]%
        {chattopadhay2018grad}
\bibfield{author}{\bibinfo{person}{Aditya Chattopadhay}, \bibinfo{person}{Anirban Sarkar}, \bibinfo{person}{Prantik Howlader}, {and} \bibinfo{person}{Vineeth~N. Balasubramanian}.} \bibinfo{year}{2018}\natexlab{}.
\newblock \showarticletitle{Grad-CAM++: Generalized Gradient-Based Visual Explanations for Deep Convolutional Networks}. In \bibinfo{booktitle}{\emph{2018 IEEE Winter Conference on Applications of Computer Vision (WACV)}}. \bibinfo{publisher}{IEEE}, \bibinfo{pages}{839--847}.
\newblock


\bibitem[Chen et~al\mbox{.}(2016)]%
        {benefit-context-1}
\bibfield{author}{\bibinfo{person}{Xi~Stephen Chen}, \bibinfo{person}{He He}, {and} \bibinfo{person}{Larry~S. Davis}.} \bibinfo{year}{2016}\natexlab{}.
\newblock \showarticletitle{Object Detection in 20 Questions}. In \bibinfo{booktitle}{\emph{2016 IEEE Winter Conference on Applications of Computer Vision (WACV)}}. \bibinfo{pages}{1--9}.
\newblock
\urldef\tempurl%
\url{https://doi.org/10.1109/WACV.2016.7477562}
\showDOI{\tempurl}


\bibitem[Clark et~al\mbox{.}(2023)]%
        {no-information-supporter}
\bibfield{author}{\bibinfo{person}{Benedict Clark}, \bibinfo{person}{Rick Wilming}, {and} \bibinfo{person}{Stefan Haufe}.} \bibinfo{year}{2023}\natexlab{}.
\newblock \showarticletitle{XAI-TRIS: Non-linear Benchmarks to Quantify ML Explanation Performance}.
\newblock \bibinfo{journal}{\emph{arXiv preprint arXiv:2306.12816}} (\bibinfo{year}{2023}).
\newblock
\urldef\tempurl%
\url{https://arxiv.org/abs/2306.12816}
\showURL{%
\tempurl}
\newblock
\shownote{arXiv preprint}.


\bibitem[Deng et~al\mbox{.}(2009)]%
        {imagenet-1k}
\bibfield{author}{\bibinfo{person}{Jia Deng}, \bibinfo{person}{Wei Dong}, \bibinfo{person}{Richard Socher}, \bibinfo{person}{Li-Jia Li}, \bibinfo{person}{Kai Li}, {and} \bibinfo{person}{Li Fei-Fei}.} \bibinfo{year}{2009}\natexlab{}.
\newblock \showarticletitle{ImageNet: A Large-Scale Hierarchical Image Database}. In \bibinfo{booktitle}{\emph{2009 IEEE Conference on Computer Vision and Pattern Recognition (CVPR)}}. \bibinfo{publisher}{IEEE}, \bibinfo{pages}{248--255}.
\newblock
\urldef\tempurl%
\url{https://doi.org/10.1109/CVPR.2009.5206848}
\showDOI{\tempurl}


\bibitem[Divvala et~al\mbox{.}(2009)]%
        {benefit-context-2}
\bibfield{author}{\bibinfo{person}{Santosh~K. Divvala}, \bibinfo{person}{Derek Hoiem}, \bibinfo{person}{James~H. Hays}, \bibinfo{person}{Alexei~A. Efros}, {and} \bibinfo{person}{Martial Hebert}.} \bibinfo{year}{2009}\natexlab{}.
\newblock \showarticletitle{An Empirical Study of Context in Object Detection}. In \bibinfo{booktitle}{\emph{2009 IEEE Conference on Computer Vision and Pattern Recognition (CVPR)}}. \bibinfo{pages}{1271--1278}.
\newblock
\urldef\tempurl%
\url{https://doi.org/10.1109/CVPR.2009.5206532}
\showDOI{\tempurl}


\bibitem[Dosovitskiy et~al\mbox{.}(2020)]%
        {vit}
\bibfield{author}{\bibinfo{person}{Alexey Dosovitskiy}, \bibinfo{person}{Lucas Beyer}, \bibinfo{person}{Alexander Kolesnikov}, \bibinfo{person}{Dirk Weissenborn}, \bibinfo{person}{Xiaohua Zhai}, \bibinfo{person}{Thomas Unterthiner}, \bibinfo{person}{Mostafa Dehghani}, \bibinfo{person}{Matthias Minderer}, \bibinfo{person}{Georg Heigold}, \bibinfo{person}{Sylvain Gelly}, {et~al\mbox{.}}} \bibinfo{year}{2020}\natexlab{}.
\newblock \showarticletitle{An Image is Worth 16x16 Words: Transformers for Image Recognition at Scale}.
\newblock \bibinfo{journal}{\emph{arXiv preprint arXiv:2010.11929}} (\bibinfo{year}{2020}).
\newblock
\urldef\tempurl%
\url{https://arxiv.org/abs/2010.11929}
\showURL{%
\tempurl}
\newblock
\shownote{arXiv preprint}.


\bibitem[Fellbaum(2010)]%
        {wordnet}
\bibfield{author}{\bibinfo{person}{Christiane Fellbaum}.} \bibinfo{year}{2010}\natexlab{}.
\newblock \showarticletitle{WordNet}.
\newblock In \bibinfo{booktitle}{\emph{Theory and Applications of Ontology: Computer Applications}}. \bibinfo{publisher}{Springer}, \bibinfo{pages}{231--243}.
\newblock


\bibitem[Galleguillos et~al\mbox{.}(2008)]%
        {benefit-context-3}
\bibfield{author}{\bibinfo{person}{Carolina Galleguillos}, \bibinfo{person}{Andrew Rabinovich}, {and} \bibinfo{person}{Serge Belongie}.} \bibinfo{year}{2008}\natexlab{}.
\newblock \showarticletitle{Object Categorization Using Co-Occurrence, Location, and Appearance}. In \bibinfo{booktitle}{\emph{2008 IEEE Conference on Computer Vision and Pattern Recognition (CVPR)}}. \bibinfo{pages}{1--8}.
\newblock
\urldef\tempurl%
\url{https://doi.org/10.1109/CVPR.2008.4587799}
\showDOI{\tempurl}


\bibitem[Gao et~al\mbox{.}(2021)]%
        {imagenet-s}
\bibfield{author}{\bibinfo{person}{Shanghua Gao}, \bibinfo{person}{Zhong-Yu Li}, \bibinfo{person}{Ming-Hsuan Yang}, \bibinfo{person}{Ming-Ming Cheng}, \bibinfo{person}{Junwei Han}, {and} \bibinfo{person}{Philip H.~S. Torr}.} \bibinfo{year}{2021}\natexlab{}.
\newblock \showarticletitle{Large-Scale Unsupervised Semantic Segmentation}.
\newblock \bibinfo{journal}{\emph{arXiv preprint arXiv:2106.03149}} (\bibinfo{year}{2021}).
\newblock
\urldef\tempurl%
\url{https://arxiv.org/abs/2106.03149}
\showURL{%
\tempurl}
\newblock
\shownote{arXiv preprint}.


\bibitem[González et~al\mbox{.}(2016)]%
        {autonomous-benefit-1}
\bibfield{author}{\bibinfo{person}{David González}, \bibinfo{person}{Joshué Pérez}, \bibinfo{person}{Vicente Milanés}, {and} \bibinfo{person}{Fawzi Nashashibi}.} \bibinfo{year}{2016}\natexlab{}.
\newblock \showarticletitle{A Review of Motion Planning Techniques for Automated Vehicles}.
\newblock \bibinfo{journal}{\emph{IEEE Transactions on Intelligent Transportation Systems}} \bibinfo{volume}{17}, \bibinfo{number}{4} (\bibinfo{year}{2016}), \bibinfo{pages}{1135--1145}.
\newblock
\urldef\tempurl%
\url{https://doi.org/10.1109/TITS.2015.2498841}
\showDOI{\tempurl}


\bibitem[He et~al\mbox{.}(2015)]%
        {dnn_passing_human}
\bibfield{author}{\bibinfo{person}{Kaiming He}, \bibinfo{person}{Xiangyu Zhang}, \bibinfo{person}{Shaoqing Ren}, {and} \bibinfo{person}{Jian Sun}.} \bibinfo{year}{2015}\natexlab{}.
\newblock \showarticletitle{Delving Deep into Rectifiers: Surpassing Human-Level Performance on ImageNet Classification}. In \bibinfo{booktitle}{\emph{Proceedings of the IEEE International Conference on Computer Vision (ICCV)}}. \bibinfo{publisher}{IEEE}.
\newblock


\bibitem[He et~al\mbox{.}(2016)]%
        {resnet}
\bibfield{author}{\bibinfo{person}{Kaiming He}, \bibinfo{person}{Xiangyu Zhang}, \bibinfo{person}{Shaoqing Ren}, {and} \bibinfo{person}{Jian Sun}.} \bibinfo{year}{2016}\natexlab{}.
\newblock \showarticletitle{Deep Residual Learning for Image Recognition}. In \bibinfo{booktitle}{\emph{Proceedings of the IEEE Conference on Computer Vision and Pattern Recognition (CVPR)}}. \bibinfo{publisher}{IEEE}, \bibinfo{pages}{770--778}.
\newblock
\urldef\tempurl%
\url{https://doi.org/10.1109/CVPR.2016.90}
\showDOI{\tempurl}


\bibitem[Hendrycks and Dietterich(2019)]%
        {imagenet-c}
\bibfield{author}{\bibinfo{person}{Dan Hendrycks} {and} \bibinfo{person}{Thomas Dietterich}.} \bibinfo{year}{2019}\natexlab{}.
\newblock \showarticletitle{Benchmarking Neural Network Robustness to Common Corruptions and Perturbations}.
\newblock \bibinfo{journal}{\emph{Proceedings of the International Conference on Learning Representations (ICLR)}} (\bibinfo{year}{2019}).
\newblock
\urldef\tempurl%
\url{https://openreview.net/forum?id=HJz6tiCqYm}
\showURL{%
\tempurl}


\bibitem[Huang et~al\mbox{.}(2020)]%
        {autonomous-driving}
\bibfield{author}{\bibinfo{person}{Xiaowei Huang}, \bibinfo{person}{Daniel Kroening}, \bibinfo{person}{Wenjie Ruan}, \bibinfo{person}{James Sharp}, \bibinfo{person}{Youcheng Sun}, \bibinfo{person}{Emese Thamo}, \bibinfo{person}{Min Wu}, {and} \bibinfo{person}{Xinping Yi}.} \bibinfo{year}{2020}\natexlab{}.
\newblock \showarticletitle{A Survey of Safety and Trustworthiness of Deep Neural Networks: Verification, Testing, Adversarial Attack and Defence, and Interpretability}.
\newblock \bibinfo{journal}{\emph{Computer Science Review}}  \bibinfo{volume}{37} (\bibinfo{year}{2020}), \bibinfo{pages}{100270}.
\newblock


\bibitem[Ilyas et~al\mbox{.}(2019)]%
        {no-information-adversarial-2}
\bibfield{author}{\bibinfo{person}{Andrew Ilyas}, \bibinfo{person}{Shibani Santurkar}, \bibinfo{person}{Dimitris Tsipras}, \bibinfo{person}{Logan Engstrom}, \bibinfo{person}{Brandon Tran}, {and} \bibinfo{person}{Aleksander Madry}.} \bibinfo{year}{2019}\natexlab{}.
\newblock \showarticletitle{Adversarial Examples Are Not Bugs, They Are Features}.
\newblock \bibinfo{journal}{\emph{Advances in Neural Information Processing Systems (NeurIPS)}}  \bibinfo{volume}{32} (\bibinfo{year}{2019}).
\newblock


\bibitem[Koh et~al\mbox{.}(2020)]%
        {CBM}
\bibfield{author}{\bibinfo{person}{Pang~Wei Koh}, \bibinfo{person}{Thao Nguyen}, \bibinfo{person}{Yew~Siang Tang}, \bibinfo{person}{Stephen Mussmann}, \bibinfo{person}{Emma Pierson}, \bibinfo{person}{Been Kim}, {and} \bibinfo{person}{Percy Liang}.} \bibinfo{year}{2020}\natexlab{}.
\newblock \showarticletitle{Concept Bottleneck Models}. In \bibinfo{booktitle}{\emph{Proceedings of the International Conference on Machine Learning (ICML)}}. PMLR, \bibinfo{pages}{5338--5348}.
\newblock
\urldef\tempurl%
\url{http://proceedings.mlr.press/v119/koh20a.html}
\showURL{%
\tempurl}


\bibitem[Lin et~al\mbox{.}(2013)]%
        {disadvantage-context-1}
\bibfield{author}{\bibinfo{person}{Dahua Lin}, \bibinfo{person}{Sanja Fidler}, {and} \bibinfo{person}{Raquel Urtasun}.} \bibinfo{year}{2013}\natexlab{}.
\newblock \showarticletitle{Holistic Scene Understanding for 3D Object Detection with RGBD Cameras}. In \bibinfo{booktitle}{\emph{2013 IEEE International Conference on Computer Vision (ICCV)}}. \bibinfo{pages}{1417--1424}.
\newblock
\urldef\tempurl%
\url{https://doi.org/10.1109/ICCV.2013.179}
\showDOI{\tempurl}


\bibitem[Litjens et~al\mbox{.}(2017)]%
        {medical}
\bibfield{author}{\bibinfo{person}{Geert Litjens}, \bibinfo{person}{Thijs Kooi}, \bibinfo{person}{Babak~Ehteshami Bejnordi}, \bibinfo{person}{Arnaud~Adiyoso Setio}, \bibinfo{person}{Francesco Ciompi}, \bibinfo{person}{Mohsen Ghafoorian}, \bibinfo{person}{Jeroen A. W.~M. van~der Laak}, \bibinfo{person}{Bram van Ginneken}, {and} \bibinfo{person}{Clara~I. S{\'a}nchez}.} \bibinfo{year}{2017}\natexlab{}.
\newblock \showarticletitle{A Survey on Deep Learning in Medical Image Analysis}.
\newblock \bibinfo{journal}{\emph{CoRR}}  \bibinfo{volume}{abs/1702.05747} (\bibinfo{year}{2017}).
\newblock
\urldef\tempurl%
\url{http://arxiv.org/abs/1702.05747}
\showURL{%
\tempurl}


\bibitem[Lundberg and Lee(2017)]%
        {SHAP}
\bibfield{author}{\bibinfo{person}{Scott Lundberg} {and} \bibinfo{person}{Su-In Lee}.} \bibinfo{year}{2017}\natexlab{}.
\newblock \bibinfo{title}{A Unified Approach to Interpreting Model Predictions}.
\newblock
\newblock
\showeprint[arxiv]{1705.07874}~[cs.AI]
\urldef\tempurl%
\url{https://arxiv.org/abs/1705.07874}
\showURL{%
\tempurl}
\newblock
\shownote{arXiv preprint}.


\bibitem[Mottaghi et~al\mbox{.}(2014)]%
        {benefit-context-4}
\bibfield{author}{\bibinfo{person}{Roozbeh Mottaghi}, \bibinfo{person}{Xianjie Chen}, \bibinfo{person}{Xiaobai Liu}, \bibinfo{person}{Nam-Gyu Cho}, \bibinfo{person}{Seong-Whan Lee}, \bibinfo{person}{Sanja Fidler}, \bibinfo{person}{Raquel Urtasun}, {and} \bibinfo{person}{Alan Yuille}.} \bibinfo{year}{2014}\natexlab{}.
\newblock \showarticletitle{The Role of Context for Object Detection and Semantic Segmentation in the Wild}. In \bibinfo{booktitle}{\emph{2014 IEEE Conference on Computer Vision and Pattern Recognition (CVPR)}}. \bibinfo{pages}{891--898}.
\newblock
\urldef\tempurl%
\url{https://doi.org/10.1109/CVPR.2014.119}
\showDOI{\tempurl}


\bibitem[Patr{\'\i}cio et~al\mbox{.}(2023)]%
        {medical-survey}
\bibfield{author}{\bibinfo{person}{Cristiano Patr{\'\i}cio}, \bibinfo{person}{Jo{\~a}o~C Neves}, {and} \bibinfo{person}{Lu{\'\i}s~F Teixeira}.} \bibinfo{year}{2023}\natexlab{}.
\newblock \showarticletitle{Explainable Deep Learning Methods in Medical Image Classification: A Survey}.
\newblock \bibinfo{journal}{\emph{Comput. Surveys}} \bibinfo{volume}{56}, \bibinfo{number}{4} (\bibinfo{year}{2023}), \bibinfo{pages}{1--41}.
\newblock


\bibitem[Ribeiro et~al\mbox{.}(2016)]%
        {LIME}
\bibfield{author}{\bibinfo{person}{Marco~Tulio Ribeiro}, \bibinfo{person}{Sameer Singh}, {and} \bibinfo{person}{Carlos Guestrin}.} \bibinfo{year}{2016}\natexlab{}.
\newblock \bibinfo{title}{"Why Should I Trust You?": Explaining the Predictions of Any Classifier}.
\newblock
\newblock
\showeprint[arxiv]{1602.04938}~[cs.LG]
\urldef\tempurl%
\url{https://arxiv.org/abs/1602.04938}
\showURL{%
\tempurl}
\newblock
\shownote{arXiv preprint}.


\bibitem[Samek et~al\mbox{.}(2017)]%
        {samek2017explainable}
\bibfield{author}{\bibinfo{person}{Wojciech Samek}, \bibinfo{person}{Thomas Wiegand}, {and} \bibinfo{person}{Klaus-Robert M{\"u}ller}.} \bibinfo{year}{2017}\natexlab{}.
\newblock \showarticletitle{Explainable Artificial Intelligence: Understanding, Visualizing and Interpreting Deep Learning Models}.
\newblock \bibinfo{journal}{\emph{arXiv preprint arXiv:1708.08296}} (\bibinfo{year}{2017}).
\newblock
\urldef\tempurl%
\url{https://arxiv.org/abs/1708.08296}
\showURL{%
\tempurl}
\newblock
\shownote{arXiv preprint}.


\bibitem[Schroff et~al\mbox{.}(2015)]%
        {face-recognition-benefit-1}
\bibfield{author}{\bibinfo{person}{Florian Schroff}, \bibinfo{person}{Dmitry Kalenichenko}, {and} \bibinfo{person}{James Philbin}.} \bibinfo{year}{2015}\natexlab{}.
\newblock \showarticletitle{FaceNet: A Unified Embedding for Face Recognition and Clustering}. In \bibinfo{booktitle}{\emph{Proceedings of the IEEE Conference on Computer Vision and Pattern Recognition (CVPR)}}. \bibinfo{publisher}{IEEE}.
\newblock


\bibitem[Selvaraju et~al\mbox{.}(2017)]%
        {selvaraju2017grad}
\bibfield{author}{\bibinfo{person}{Ramprasaath~R. Selvaraju}, \bibinfo{person}{Michael Cogswell}, \bibinfo{person}{Abhishek Das}, \bibinfo{person}{Ramakrishna Vedantam}, \bibinfo{person}{Devi Parikh}, {and} \bibinfo{person}{Dhruv Batra}.} \bibinfo{year}{2017}\natexlab{}.
\newblock \showarticletitle{Grad-CAM: Visual Explanations from Deep Networks via Gradient-Based Localization}. In \bibinfo{booktitle}{\emph{Proceedings of the IEEE International Conference on Computer Vision (ICCV)}}. \bibinfo{publisher}{IEEE}, \bibinfo{pages}{618--626}.
\newblock
\urldef\tempurl%
\url{https://openaccess.thecvf.com/content_iccv_2017/html/Selvaraju_Grad-CAM_Visual_Explanations_ICCV_2017_paper.html}
\showURL{%
\tempurl}


\bibitem[Simonyan and Zisserman(2014)]%
        {vgg}
\bibfield{author}{\bibinfo{person}{Karen Simonyan} {and} \bibinfo{person}{Andrew Zisserman}.} \bibinfo{year}{2014}\natexlab{}.
\newblock \showarticletitle{Very Deep Convolutional Networks for Large-Scale Image Recognition}.
\newblock \bibinfo{journal}{\emph{arXiv preprint arXiv:1409.1556}} (\bibinfo{year}{2014}).
\newblock
\urldef\tempurl%
\url{https://arxiv.org/abs/1409.1556}
\showURL{%
\tempurl}
\newblock
\shownote{arXiv preprint}.


\bibitem[Springenberg et~al\mbox{.}(2014)]%
        {springenberg2014striving}
\bibfield{author}{\bibinfo{person}{Jost~Tobias Springenberg}, \bibinfo{person}{Alexey Dosovitskiy}, \bibinfo{person}{Thomas Brox}, {and} \bibinfo{person}{Martin Riedmiller}.} \bibinfo{year}{2014}\natexlab{}.
\newblock \showarticletitle{Striving for Simplicity: The All Convolutional Net}.
\newblock \bibinfo{journal}{\emph{arXiv preprint arXiv:1412.6806}} (\bibinfo{year}{2014}).
\newblock
\urldef\tempurl%
\url{https://arxiv.org/abs/1412.6806}
\showURL{%
\tempurl}
\newblock
\shownote{arXiv preprint}.


\bibitem[Srinivas and Fleuret(2019)]%
        {srinivas2019full}
\bibfield{author}{\bibinfo{person}{Suraj Srinivas} {and} \bibinfo{person}{Fran{\c{c}}ois Fleuret}.} \bibinfo{year}{2019}\natexlab{}.
\newblock \showarticletitle{Full-Gradient Representation for Neural Network Visualization}.
\newblock \bibinfo{journal}{\emph{Advances in Neural Information Processing Systems (NeurIPS)}}  \bibinfo{volume}{32} (\bibinfo{year}{2019}).
\newblock


\bibitem[Szegedy et~al\mbox{.}(2015)]%
        {inception-net}
\bibfield{author}{\bibinfo{person}{Christian Szegedy}, \bibinfo{person}{Wei Liu}, \bibinfo{person}{Yangqing Jia}, \bibinfo{person}{Pierre Sermanet}, \bibinfo{person}{Scott Reed}, \bibinfo{person}{Dragomir Anguelov}, \bibinfo{person}{Dumitru Erhan}, \bibinfo{person}{Vincent Vanhoucke}, {and} \bibinfo{person}{Andrew Rabinovich}.} \bibinfo{year}{2015}\natexlab{}.
\newblock \showarticletitle{Going Deeper with Convolutions}. In \bibinfo{booktitle}{\emph{Proceedings of the IEEE Conference on Computer Vision and Pattern Recognition (CVPR)}}. \bibinfo{pages}{1--9}.
\newblock


\bibitem[Taesiri et~al\mbox{.}(2023)]%
        {zoom-required}
\bibfield{author}{\bibinfo{person}{Mohammad~Reza Taesiri}, \bibinfo{person}{Giang Nguyen}, \bibinfo{person}{Sarra Habchi}, \bibinfo{person}{Cor-Paul Bezemer}, {and} \bibinfo{person}{Anh Nguyen}.} \bibinfo{year}{2023}\natexlab{}.
\newblock \showarticletitle{Zoom is What You Need: An Empirical Study of the Power of Zoom and Spatial Biases in Image Classification}.
\newblock \bibinfo{journal}{\emph{arXiv preprint arXiv:2304.05538}} (\bibinfo{year}{2023}).
\newblock
\urldef\tempurl%
\url{https://arxiv.org/abs/2304.05538}
\showURL{%
\tempurl}
\newblock
\shownote{arXiv preprint}.


\bibitem[Tan and Le(2019)]%
        {efficientnet}
\bibfield{author}{\bibinfo{person}{Mingxing Tan} {and} \bibinfo{person}{Quoc Le}.} \bibinfo{year}{2019}\natexlab{}.
\newblock \showarticletitle{EfficientNet: Rethinking Model Scaling for Convolutional Neural Networks}. In \bibinfo{booktitle}{\emph{Proceedings of the International Conference on Machine Learning (ICML)}}. PMLR, \bibinfo{pages}{6105--6114}.
\newblock
\urldef\tempurl%
\url{http://proceedings.mlr.press/v97/tan19a.html}
\showURL{%
\tempurl}


\bibitem[Tian et~al\mbox{.}(2023)]%
        {tian2023learning}
\bibfield{author}{\bibinfo{person}{Zhuotao Tian}, \bibinfo{person}{Jiequan Cui}, \bibinfo{person}{Li Jiang}, \bibinfo{person}{Xiaojuan Qi}, \bibinfo{person}{Xin Lai}, \bibinfo{person}{Yixin Chen}, \bibinfo{person}{Shu Liu}, {and} \bibinfo{person}{Jiaya Jia}.} \bibinfo{year}{2023}\natexlab{}.
\newblock \bibinfo{title}{Learning Context-Aware Classifier for Semantic Segmentation}.
\newblock
\newblock
\showeprint[arxiv]{2303.11633}~[cs.CV]
\newblock
\shownote{arXiv preprint}.


\bibitem[Vaswani et~al\mbox{.}(2017)]%
        {transformers}
\bibfield{author}{\bibinfo{person}{Ashish Vaswani}, \bibinfo{person}{Noam Shazeer}, \bibinfo{person}{Niki Parmar}, \bibinfo{person}{Jakob Uszkoreit}, \bibinfo{person}{Llion Jones}, \bibinfo{person}{Aidan~N Gomez}, \bibinfo{person}{{\L}ukasz Kaiser}, {and} \bibinfo{person}{Illia Polosukhin}.} \bibinfo{year}{2017}\natexlab{}.
\newblock \showarticletitle{Attention is All You Need}.
\newblock \bibinfo{journal}{\emph{Advances in Neural Information Processing Systems (NeurIPS)}}  \bibinfo{volume}{30} (\bibinfo{year}{2017}).
\newblock


\bibitem[Wang et~al\mbox{.}(2020)]%
        {wang2020score}
\bibfield{author}{\bibinfo{person}{Haofan Wang}, \bibinfo{person}{Zifan Wang}, \bibinfo{person}{Mengnan Du}, \bibinfo{person}{Fan Yang}, \bibinfo{person}{Zijian Zhang}, \bibinfo{person}{Sirui Ding}, \bibinfo{person}{Piotr Mardziel}, {and} \bibinfo{person}{Xia Hu}.} \bibinfo{year}{2020}\natexlab{}.
\newblock \showarticletitle{Score-CAM: Score-Weighted Visual Explanations for Convolutional Neural Networks}. In \bibinfo{booktitle}{\emph{Proceedings of the IEEE/CVF Conference on Computer Vision and Pattern Recognition Workshops (CVPRW)}}. \bibinfo{publisher}{IEEE}, \bibinfo{pages}{24--25}.
\newblock


\bibitem[Wang et~al\mbox{.}(2016)]%
        {motion-benefit-1}
\bibfield{author}{\bibinfo{person}{Limin Wang}, \bibinfo{person}{Yuanjun Xiong}, \bibinfo{person}{Zhe Wang}, \bibinfo{person}{Yu Qiao}, \bibinfo{person}{Dahua Lin}, \bibinfo{person}{Xiaoou Tang}, {and} \bibinfo{person}{Luc Van~Gool}.} \bibinfo{year}{2016}\natexlab{}.
\newblock \showarticletitle{Temporal Segment Networks: Towards Good Practices for Deep Action Recognition}. In \bibinfo{booktitle}{\emph{European Conference on Computer Vision (ECCV)}}. \bibinfo{publisher}{Springer}, \bibinfo{pages}{20--36}.
\newblock


\bibitem[Wang et~al\mbox{.}(2023)]%
        {frequency-shortcuts}
\bibfield{author}{\bibinfo{person}{Shunxin Wang}, \bibinfo{person}{Raymond Veldhuis}, \bibinfo{person}{Christoph Brune}, {and} \bibinfo{person}{Nicola Strisciuglio}.} \bibinfo{year}{2023}\natexlab{}.
\newblock \showarticletitle{What Do Neural Networks Learn in Image Classification? A Frequency Shortcut Perspective}. In \bibinfo{booktitle}{\emph{Proceedings of the IEEE/CVF International Conference on Computer Vision (ICCV)}}. \bibinfo{pages}{1433--1442}.
\newblock


\bibitem[Wang and Zhu(2023)]%
        {benefit-context-5}
\bibfield{author}{\bibinfo{person}{Xuan Wang} {and} \bibinfo{person}{Zhigang Zhu}.} \bibinfo{year}{2023}\natexlab{}.
\newblock \showarticletitle{Context Understanding in Computer Vision: A Survey}.
\newblock \bibinfo{journal}{\emph{Computer Vision and Image Understanding}}  \bibinfo{volume}{229} (\bibinfo{date}{March} \bibinfo{year}{2023}), \bibinfo{pages}{103646}.
\newblock
\urldef\tempurl%
\url{https://doi.org/10.1016/j.cviu.2023.103646}
\showDOI{\tempurl}


\bibitem[Wightman(2019)]%
        {timm}
\bibfield{author}{\bibinfo{person}{Ross Wightman}.} \bibinfo{year}{2019}\natexlab{}.
\newblock \bibinfo{title}{PyTorch Image Models}.
\newblock \bibinfo{howpublished}{GitHub repository}.
\newblock
\urldef\tempurl%
\url{https://doi.org/10.5281/zenodo.4414861}
\showDOI{\tempurl}


\bibitem[Wilming et~al\mbox{.}(2022)]%
        {no-information-supporter-backup}
\bibfield{author}{\bibinfo{person}{Rick Wilming}, \bibinfo{person}{C{\'e}line Budding}, \bibinfo{person}{Klaus-Robert M{\"u}ller}, {and} \bibinfo{person}{Stefan Haufe}.} \bibinfo{year}{2022}\natexlab{}.
\newblock \showarticletitle{Scrutinizing XAI using Linear Ground-Truth Data with Suppressor Variables}.
\newblock \bibinfo{journal}{\emph{Machine Learning}} \bibinfo{volume}{111}, \bibinfo{number}{5} (\bibinfo{year}{2022}), \bibinfo{pages}{1903--1923}.
\newblock
\urldef\tempurl%
\url{https://doi.org/10.1007/s10994-022-06171-0}
\showDOI{\tempurl}


\bibitem[Xiao et~al\mbox{.}(2021)]%
        {xiao2021noise}
\bibfield{author}{\bibinfo{person}{Kai~Yuanqing Xiao}, \bibinfo{person}{Logan Engstrom}, \bibinfo{person}{Andrew Ilyas}, {and} \bibinfo{person}{Aleksander Madry}.} \bibinfo{year}{2021}\natexlab{}.
\newblock \showarticletitle{Noise or Signal: The Role of Image Backgrounds in Object Recognition}. In \bibinfo{booktitle}{\emph{Proceedings of the International Conference on Learning Representations (ICLR)}}.
\newblock
\urldef\tempurl%
\url{https://openreview.net/forum?id=gl3D-xY7wLq}
\showURL{%
\tempurl}


\bibitem[Yang and Kim(2019)]%
        {paper-4-google_brain}
\bibfield{author}{\bibinfo{person}{Mengjiao Yang} {and} \bibinfo{person}{Been Kim}.} \bibinfo{year}{2019}\natexlab{}.
\newblock \showarticletitle{Benchmarking Attribution Methods with Relative Feature Importance}.
\newblock \bibinfo{journal}{\emph{arXiv preprint arXiv:1907.09701}} (\bibinfo{year}{2019}).
\newblock
\urldef\tempurl%
\url{https://arxiv.org/abs/1907.09701}
\showURL{%
\tempurl}
\newblock
\shownote{arXiv preprint}.


\bibitem[Yao et~al\mbox{.}(2012)]%
        {disadvantage-context-2}
\bibfield{author}{\bibinfo{person}{Jian Yao}, \bibinfo{person}{Sanja Fidler}, {and} \bibinfo{person}{Raquel Urtasun}.} \bibinfo{year}{2012}\natexlab{}.
\newblock \showarticletitle{Describing the Scene as a Whole: Joint Object Detection, Scene Classification, and Semantic Segmentation}. In \bibinfo{booktitle}{\emph{2012 IEEE Conference on Computer Vision and Pattern Recognition (CVPR)}}. \bibinfo{pages}{702--709}.
\newblock
\urldef\tempurl%
\url{https://doi.org/10.1109/CVPR.2012.6247739}
\showDOI{\tempurl}


\bibitem[Yu et~al\mbox{.}(2016)]%
        {context-selective}
\bibfield{author}{\bibinfo{person}{Ruichi Yu}, \bibinfo{person}{Xi Chen}, \bibinfo{person}{Vlad~I. Morariu}, {and} \bibinfo{person}{Larry~S. Davis}.} \bibinfo{year}{2016}\natexlab{}.
\newblock \showarticletitle{The Role of Context Selection in Object Detection}.
\newblock \bibinfo{journal}{\emph{arXiv preprint arXiv:1609.02948}} (\bibinfo{year}{2016}).
\newblock
\urldef\tempurl%
\url{https://arxiv.org/abs/1609.02948}
\showURL{%
\tempurl}
\newblock
\shownote{arXiv preprint}.


\bibitem[Zhang(2019)]%
        {pixelate-explain}
\bibfield{author}{\bibinfo{person}{Richard Zhang}.} \bibinfo{year}{2019}\natexlab{}.
\newblock \showarticletitle{Making Convolutional Networks Shift-Invariant Again}. In \bibinfo{booktitle}{\emph{Proceedings of the 36th International Conference on Machine Learning (ICML)}} \emph{(\bibinfo{series}{Proceedings of Machine Learning Research}, Vol.~\bibinfo{volume}{97})}. \bibinfo{publisher}{PMLR}, \bibinfo{pages}{7324--7334}.
\newblock
\urldef\tempurl%
\url{http://proceedings.mlr.press/v97/zhang19a.html}
\showURL{%
\tempurl}


\bibitem[Zhou et~al\mbox{.}(2022)]%
        {feature-attribution-closest-to-us}
\bibfield{author}{\bibinfo{person}{Yilun Zhou}, \bibinfo{person}{Serena Booth}, \bibinfo{person}{Marco~Tulio Ribeiro}, {and} \bibinfo{person}{Julie Shah}.} \bibinfo{year}{2022}\natexlab{}.
\newblock \showarticletitle{Do Feature Attribution Methods Correctly Attribute Features?}. In \bibinfo{booktitle}{\emph{Proceedings of the AAAI Conference on Artificial Intelligence}}, Vol.~\bibinfo{volume}{36}. \bibinfo{pages}{9623--9633}.
\newblock


\bibitem[Zhu et~al\mbox{.}(2016)]%
        {object_with_or_without_object}
\bibfield{author}{\bibinfo{person}{Zhuotun Zhu}, \bibinfo{person}{Lingxi Xie}, {and} \bibinfo{person}{Alan~L. Yuille}.} \bibinfo{year}{2016}\natexlab{}.
\newblock \showarticletitle{Object Recognition with and without Objects}.
\newblock \bibinfo{journal}{\emph{arXiv preprint arXiv:1611.06596}} (\bibinfo{year}{2016}).
\newblock
\urldef\tempurl%
\url{https://arxiv.org/abs/1611.06596}
\showURL{%
\tempurl}
\newblock
\shownote{arXiv preprint}.


\end{thebibliography}
\bibliographystyle{ACM-Reference-Format}

\end{document}